\documentclass[letterpaper, 10 pt, conference]{ieeeconf}  

\IEEEoverridecommandlockouts                              

\usepackage{graphicx}
\usepackage{booktabs}
\usepackage{subcaption}
\usepackage{mathtools}
\usepackage{amsmath}
\usepackage{multicol}
\usepackage{multirow}
\usepackage{soul}
\usepackage[dvipsnames]{xcolor}

\usepackage{cite}
\usepackage{hyperref}
\hypersetup{
    colorlinks=true,
    linkcolor=blue,
    filecolor=magenta,      
    urlcolor=cyan,
    pdftitle={Overleaf Example},
    pdfpagemode=FullScreen,
    }
\hypersetup{hidelinks}

\title{\LARGE \bf
DRIFT: Deep Reinforcement Learning for Intelligent Floating Platforms Trajectories}

\author{Matteo El-Hariry$^*$$^{1}$, Antoine Richard$^*$$^{1}$, Vivek Muralidharan$^{1}$, Matthieu Geist$^{2}$, Miguel Olivares-Mendez$^{1}$
\thanks{$^{1}$Space Robotics (SpaceR) Research Group, SnT, University of Luxembourg, matteo.elhariry@uni.lu }%
\thanks{$^{2}$Matthieu Geist is with Cohere}%
}

\begin{document}

\bstctlcite{BSTcontrol}  

\maketitle
\thispagestyle{empty}
\pagestyle{empty}


\begin{abstract}

This investigation introduces a novel deep reinforcement learning-based suite to control floating platforms in both simulated and real-world environments.
Floating platforms serve as versatile test-beds to emulate microgravity environments on Earth, useful to test autonomous navigation systems for space applications.
Our approach addresses the system and environmental uncertainties in controlling such platforms by training policies capable of precise maneuvers amid dynamic and unpredictable conditions.
Leveraging Deep Reinforcement Learning (DRL) techniques, our suite achieves robustness, adaptability, and good transferability from simulation to reality.
Our deep reinforcement learning framework provides advantages such as fast training times, large-scale testing capabilities, rich visualization options, and ROS bindings for integration with real-world robotic systems.
Being open access, our suite serves as a comprehensive platform for practitioners who want to replicate similar research in their own simulated environments and labs.

\end{abstract}


\section{INTRODUCTION}

Across the globe, there has been an exponential growth in the adoption of small satellites, including cubesats~\cite{small_sats_overview,curzi2020constellations}.
Thanks to their cost-effectiveness, they are now extensively used for both commercial and scientific purposes.
Consequently, several countries and their space administrations have actively invested in advancing small satellite technology over the last few decades.
The surge in space missions is creating a growing demand to test and validate the flight software and hardware on the ground prior to employing them in space.
These experiments aim to enhance the historically low success rates of missions in space~\cite{gnc_future_missions}.

Improving the reliability and autonomy of the motion systems plays a key role in boosting mission success.
Currently, the primary approach to enhancing autonomous navigation and control of such systems involves conducting performance tests.
These tests help understand essential parameters and their relationships within the control scheme involving different sensors and actuators.
This knowledge is pivotal to the design and operation of these systems, contributing significantly to mission success rates~\cite{small_sats_overview}.
To emulate free-floating and free-flying satellite motion, a common solution is to use floating platforms: a rigid structure floating on top of an extremely flat and smooth surface using air bearings~\cite{fp_rybus2016planar}.
This allows 2D motions with very low friction, effectively replicating space-like conditions in a plane.

The traditional methods of aerospace vehicle trajectory planning primarily rely on optimal control techniques~\cite{gnc_future_missions}, which involve the derivation of open-loop control solutions based on system models and predefined objectives.
These approaches are intricate and often necessitate specialized optimization software to find flight paths adhering to various constraints.
However, as aerospace vehicles regularly encounter state disturbances and uncertainties, there is a need for more robust and adaptable control strategies.
In response to this challenge, this paper introduces a novel approach that harnesses the power of Deep Reinforcement Learning (DRL) to control a floating platform (FP) within a 2D environment.
The utilization of DRL offers an alternative to the conventional deterministic and expert-driven control methods prevalent in aerospace trajectory planning.
DRL involves a goal-oriented agent that interacts with its environment, learning control policy approximations.
Importantly, this learning process enables the agent to handle stochastic events in the environments by exploring the state-control space using reward signals.

\begin{figure}[t!] 
\vspace{6pt}
    \centering  
    \includegraphics[width=\linewidth]{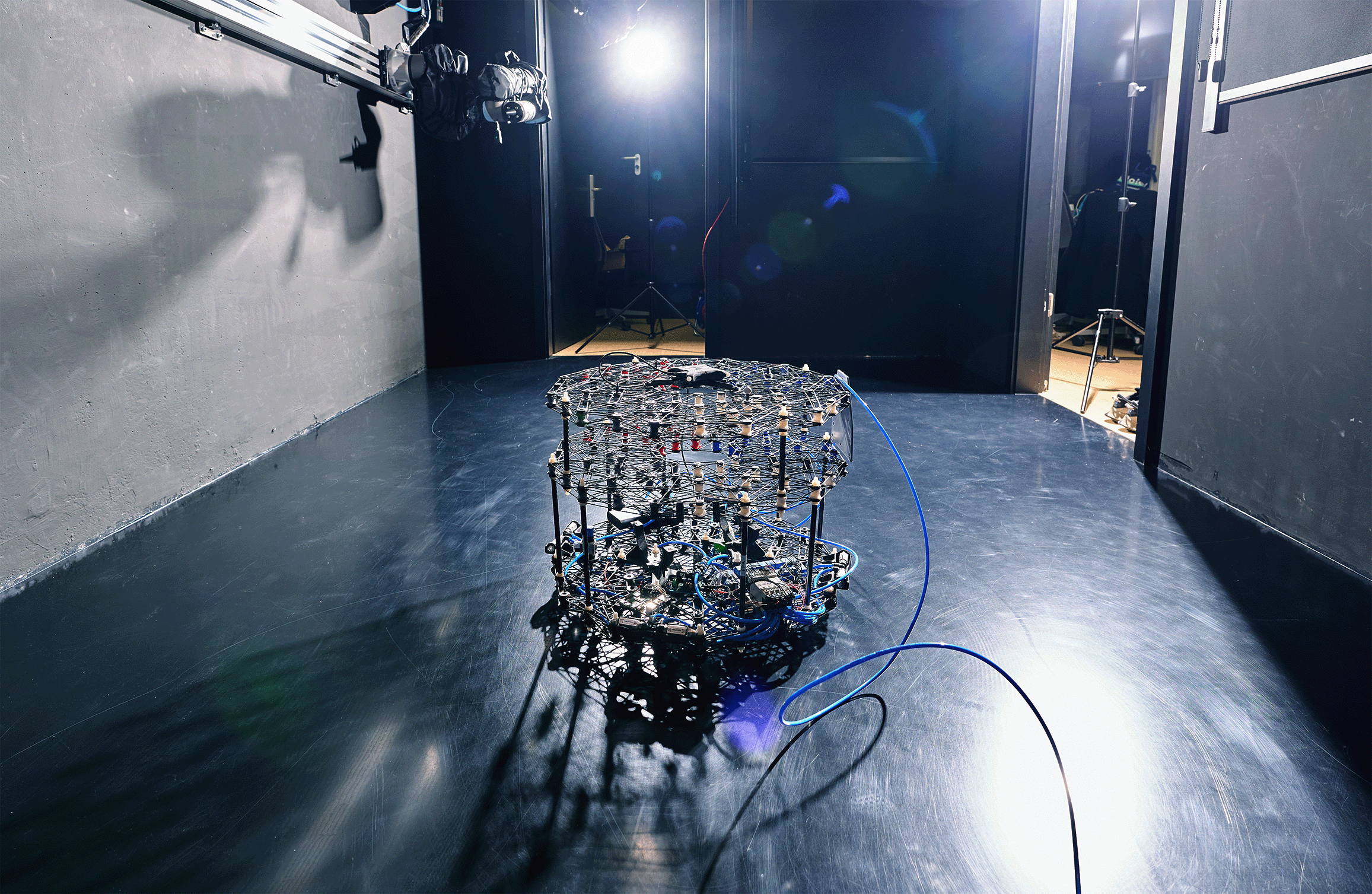}
    \caption{Floating platform in ZeroG Laboratory.}
    \label{fig:fp_ref_frames}
    \vspace{-5pt}
\end{figure}

Our main contributions lie in three key areas.
First, we enhance RANS~\cite{rans-astra23}, a simulator previously developed by our team, to accommodate more complex tasks and a diverse range of environment randomization profiles.
Second, we demonstrate the high-performance capabilities of the Proximal Policy Optimization (PPO)~\cite{schulman2017proximal} algorithm in both simulated and real scenarios.
We evaluate its effectiveness by completing two distinct tasks: navigating to a specific position and orientation, and tracking a target velocity.
Finally, we conduct a comprehensive comparison between the PPO-based approach and traditional optimal control algorithms, such as the Linear Quadratic Regulator (LQR)~\cite{khosravi2024design}, showcasing the benefits and drabacks of the two methods under different environmental conditions.



\section{Related Work}

\subsection{Floating Platforms control}
Floating platforms are systems with air bearings attached to their lower surface. These bearings release pressurized air creating a thin film to levitate the platform; thereby counterbalancing its weight to produce a microgravity effect (in-plane components of gravity on the main body are negligible), thus emulating the friction-less and weightless environment of orbital spaceflight.

Recently, many research labs and organizations have focused on developing air bearings-based simulators with 3-DoF robotic systems~\cite{fp_banerjee2022slider, fp_huang2022air_bearing_testbed, fp_sabatini2017pinocchio, fp_santaguida2023rendezvous_capture, fp_vyas2022_traj_optoation, fp_kwok2018design, fp_nieto2021formation_maneuvers, fp_rybus2016planar, fp_wapman2021jet}, making them the most popular testing facility to emulate microgravity on Earth. 

To emulate mission scenarios for autonomous spacecraft tracking, servicing, rendezvous, and capture of a free-floating target, several works have further improved these platforms with 3-DoF robotic manipulators~\cite{fp_sabatini2017pinocchio, fp_santaguida2023rendezvous_capture, fp_rl_cao2023reinforcement}. A typical approach for robotic arms mounted on air bearings platforms is to decouple the platform and the arm maneuvers. Sabatini et al.~\cite{fp_sabatini2017pinocchio} focus on obtaining a coordinated maneuver in which the end effector moves thanks to the platform motion, hence optimizing fuel efficiency. They provide results both in simulation and on a real FP.

When focusing solely on the maneuvers of free-floating platforms, noteworthy developments have emerged. For instance, an innovative 3D-printed platform named ``Slider" has been introduced~\cite{fp_banerjee2022slider}. Slider, equipped with eight thrusters, can be precisely controlled through either motion in one of the 4 cardinal directions or 2 rotations. Furthermore, \cite{fp_huang2022air_bearing_testbed} presents an extensive characterization of air bearings platforms while introducing a vision-based navigation system that takes into account the vibrations caused by the thrusters.

\subsection{Deep Reinforcement Learning for thrust-based control}
Over the past few years, there has been a flourishing utilization of ML and RL applied to space Guidance Navigation and Control (GNC) problems. These applications encompass a wide range of tasks, including: planetary landing~\cite{gaudet2020deep}; path planning for lunar or asteroid hopping rovers~\cite{yu2021path_planning, tanaka2021hopper}; spacecraft orbit control within unknown gravitational fields~\cite{Willis2016ReinforcementLF}; and spacecraft map generation during orbits around small celestial bodies~\cite{chan2019mapping}. However, it is worth noting that most research employing deep reinforcement learning for aerospace control tasks shows numerical simulation results only.   

To the best of our knowledge, only two prior works~\cite{fp_rl_hovell2020rl_spacecraft, athauda2023intelligent}, used reinforcement learning to guide both simulated and real-world floating platforms. In \cite{fp_rl_hovell2020rl_spacecraft}, the authors combine DRL as a guidance policy whose trajectories are fed to a conventional controller to track. This work provides guidance techniques that successfully output velocity signals for the simulation and the experimental facility, achieving comparable performance to that observed during training. In \cite{athauda2023intelligent} the main focus is on reaching a specific target while avoiding static or dynamic obstacles.

Our work distinguishes itself from~\cite{fp_rl_hovell2020rl_spacecraft} and~\cite{athauda2023intelligent} in several key aspects. Firstly, we introduce a DRL agent capable of directly controlling the output thrust of the floating platform, eliminating the need for a separate trajectory tracker scheme. Secondly, our primary focus lies in delivering a comprehensive framework for training, assessing, and bench-marking DRL agents and optimal control methods across a spectrum of environmental conditions. These conditions represent a significant source of complexity when deploying AI agents on real systems, especially within the demanding space environment.
\looseness=-1

\section{Methods}

\subsection{Problem Formulation}

\begin{figure}[t!] 
    \centering  
    \includegraphics[width=.8\linewidth]{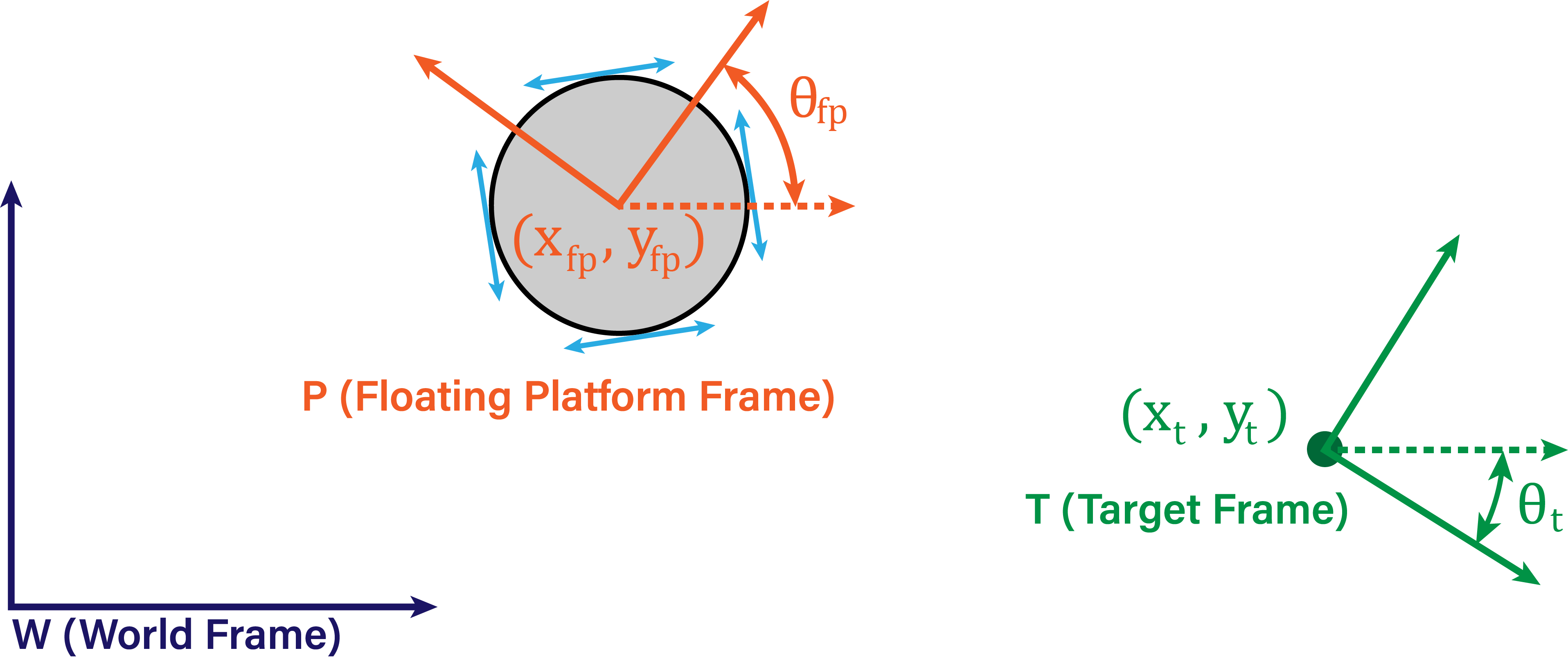}
    \caption{Floating platform and target in global reference frame.}
    \label{fig:fp_ref_frames}
    \vspace{-18pt}
\end{figure}

In this paper, we approach the task of guiding a FP's maneuvers as a sequential decision-making problem. To facilitate our investigation and demonstrate the practical applicability of our proposed techniques—from sim to real-world scenarios, we simplify the complex orbital dynamics into a two-dimensional kinematic model. As illustrated in Figure~\ref{fig:fp_ref_frames}, we use a global reference frame (denoted $W$). This allows for consistent and absolute measurements of the position and heading errors. The framework also allows for the use of local coordinates whenever considered convenient. 

Within this framework the control policy must learn the optimal sequence of actions by observing state transitions, thereby minimizing the task-specific error. 
We define the different tasks as: \textit{(i) Go to pose}, starting from a random initial position in the plane, reach the given pose (position and orientation $\theta$); \textit{(ii) Track velocity}, track the given velocity vector, which can in turn be used to follow a trajectory. 
\looseness=-1

For both tasks the control policy is required to minimize the error metrics derived from the current state observations of the floating platform and the target.
Regarding the ``go to pose'' task, the positional error is defined as the Euclidean distance between the FP's current position, $p_{fp} = (x_{fp}, y_{fp})$, and the target position, $p_{t} = (x_{t}, y_{t})$, Eq.~\eqref{eq:pos_error}, while the heading error is calculated based on the difference between the platform's current orientation $\theta_{fp}$ and the target heading $\theta_t$, Eq.~\eqref{eq:heading_error}: 
\looseness=-1
\begin{align}
    e_{p} &= \|\mathbf{p}_{\text{t}} - \mathbf{p}_{\text{fp}}\|_{\scriptscriptstyle 2}
    \label{eq:pos_error}
    \\
    e_{\theta} &= \arctan2\left(\sin(\theta_{\text{t}} - \theta_{\text{fp}}), \cos(\theta_{\text{t}} - \theta_{\text{fp}})\right)
    \label{eq:heading_error}
\end{align}




For the ``track velocity'' task the angular and linear velocity errors (\( \mathbf{e}_{v}, \mathbf{e}_{\omega} \)) are determined by subtracting the FP's current velocities ($\mathbf{v}_{fp}$) from the target velocities ($\mathbf{v}_t$), Eq.~\eqref{eq:lin_vel_error} and~\eqref{eq:ang_vel_error}. 
\begin{align}
    \label{eq:lin_vel_error}
    \mathbf{e}_{v} &= \mathbf{v}_{\text{t}} - \mathbf{v}_{\text{fp}}
    \\
    \label{eq:ang_vel_error}
    \mathbf{e}_{\omega} &= \boldsymbol{\omega}_{\text{t}} - \boldsymbol{\omega}_{\text{fp}}
\end{align}
    


In our study, we focus on a lightweight floating platform (FP) developed at the University of Luxembourg~\cite{fp_spacer}. The system is defined by a 10-dimensional state space, Eq.~\ref{eq:state}. At each discrete time step \(t\), the state variables include the FP's heading (\(\theta\)), its linear velocities (\(v_x\) and \(v_y\)), angular velocity (\(\omega_z\)), a task flag (\(\text{f}\)) indicating the current task, and four additional variables (\(\text{d}_{1-4}\)) representing task-specific data such as distances to the target position and heading:
%
\begin{equation}\label{eq:state}
    s_t = (\cos(\theta), \sin(\theta), v_x, v_y, \omega_z, \text{f}, \text{d}_1, \text{d}_2, \text{d}_3, \text{d}_4)^\top.
\end{equation}

Task-specific data, written \(\text{d}_{1-4}\), is detailed in Table~\ref{tab:task_data}, where \(\Delta\) denotes the vector norm distance between the variables (such as position, velocity, or angle) and their respective target values. This configuration of the observation space is intentionally designed to facilitate the future extension of this work to learn policies capable of handling multiple tasks simultaneously.
\looseness=-1

For the control of the platform, our agents use an 8-dimensional action space that corresponds to a binary activation of 8 ``on-off thrusters''. These share the same pressure line, such that, at every step of the control loop, the maximum force generated by each thruster is $\frac{1}{n}$~N where $n$ is the number of active thrusters. Simply put, if only one thruster is turned on, it will output 1 Newton, if 2 thrusters are activated they generate 0.5~N each, etc.

\begin{table}[t]
	\centering
    \caption{State task-specific data.}
    \label{tab:task_data}
	\begin{tabular}{cccccc}
		\hline
        \textbf{Task}            & \textbf{$\text{f}$} & \textbf{$\text{d}_1$} & \textbf{$\text{d}_2$} & \textbf{$\text{d}_3$}         & \textbf{$\text{d}_4$ }        \\ \hline
        Go to pose      & 1           & $\Delta x$    & $\Delta y$    & $\cos(\Delta \theta)$ & $\sin(\Delta \theta)$ \\
        Track velocity  & 2           & $\Delta v_x$  & $\Delta v_y$  & -                     & -                     \\ \hline
	\end{tabular}
 \vspace{-15pt}
\end{table}

 To guide the optimization process for the control policies, an exponential reward structure was adopted, as after empirical evaluation it was found to yield faster and more accurate convergence. In particular, Eq.~\eqref{eq:reward1} for the ``go to pose'' task and Eq.~\eqref{eq:reward2} for the ``track velocity'' task were used:
 \begin{align}
     \label{eq:reward1}
    R_{po} &= \exp\left(-\frac{e_p}{0.25}\right) \cdot S_p + \exp\left(-\frac{e_{\theta}}{0.25}\right) \cdot S_{\theta} - p
    \\
    \label{eq:reward2}
    R_v &= \exp\left(-\frac{e_v}{0.25}\right) \cdot S_p + \exp\left(-\frac{e_{\omega}}{0.25}\right)  \cdot S_{\theta} - p
 \end{align}


In this context, errors are quantified as the norm distance from the specified targets, with $e_v$ denoting the linear velocity error, and $e_p$ and $e_{\theta}$ representing the errors in position and orientation, respectively. Scaling coefficients $S_p$ and $S_{\theta}$, which adjust the impact of position and orientation errors, were both set to 0.5 in our experiments. Additionally, $p$ sums up to three penalties ($p_{act}$, $p_{vel}$, $p_{\omega}$) designed to discourage excessive thruster activation or reaching states with elevated linear and angular velocities. Our experimentation with various penalty configurations led us to adopt a penalty for thruster activation, Eq. \eqref{eq:act_penalty} as well as excessive angular velocities, Eq.~\eqref{eq:vel_penalty}. Here, $T$ stands for an indicator function reflecting the on-off states of the thrusters.
\begin{align}
    \label{eq:act_penalty}
    p_{act} &= 0.3 \sum_{i=1}^{8} T_i
    \\
    \label{eq:vel_penalty}
    p_{\omega} &= 0.15 \max(0, |\omega_z| - 1)
\end{align}
\subsection{Simulation}
Building upon our prior simulator RANS~\cite{rans-astra23}, we introduce enhancements to enable the platform to perform more complex tasks.
RANS leverage Nvidia's IsaacSim, specifically relying on OmniIsaacGym~\cite{makoviychuk2021isaac}, a versatile simulator, capable of concurrently running thousands of environments. In the original RANS framework, only nominal system and environmental conditions were present. This hindered the ability of the agents to adapt to non-ideal conditions, which are usually common when using the real FP systems. To mitigate this gap, we introduce RANS v2.0 which includes the following extensions:
(1) parameterized rewards and penalties, to allow easy fine-tuning of the control policies; (2) analogue kinematic model in Mujoco~\cite{mujoco},
to allow easy evaluation of both traditional and RL-based controllers in a non-Torch depended environment; (3) disturbance generation module, that allows the injection of: (a) Action Noise (AN): a random disturbance force of $\pm$ $an$ N applied to every thruster; (b) Velocity Noise (VN): $\pm$ $vn$ $\text{m/s}$ added to the state velocities; (c)  Uneven Floor (UF): $uf$ N of force, added to simulate the floor unevenness, applied to the FP body throughout the episode, either with a constant direction or through a sinusoidal generated direction; (d) Torque Disturbance (TD): $td$ Nm of torque applied to the body's center of mass; (e) Random Thrusters Failure (RTF): a zeroing mask over the output actions to simulate one or multiple thruster failures which remains the same throughout the episode. 
\looseness=-1


RANS v2.0, requires 30 minutes to train an agent on an RTX 4090.
Achieving a throughput of more than 40,000 steps per second with all disturbances enabled, which is very close to its previous version.
Furthermore, it enables large-scale testing by swiftly evaluating thousands of initial conditions in seconds.
It offers rich visualization options, including metric tracking during training through the WandB API~\cite{wandb}, 
and comprehensive evaluation metrics presented through tables and plots.
The library uses the OpenAI Gym~\cite{brockman2016openai}
format to define the RL loop, including the standard normalization of the observation space.
Additionally, the integration of a ROS interface enhances the versatility of our framework, allowing easy integration and deployment of the control policies within real-world robotic systems.


\begin{figure*}[htp] 
    \centering
    \includegraphics[width=\textwidth]{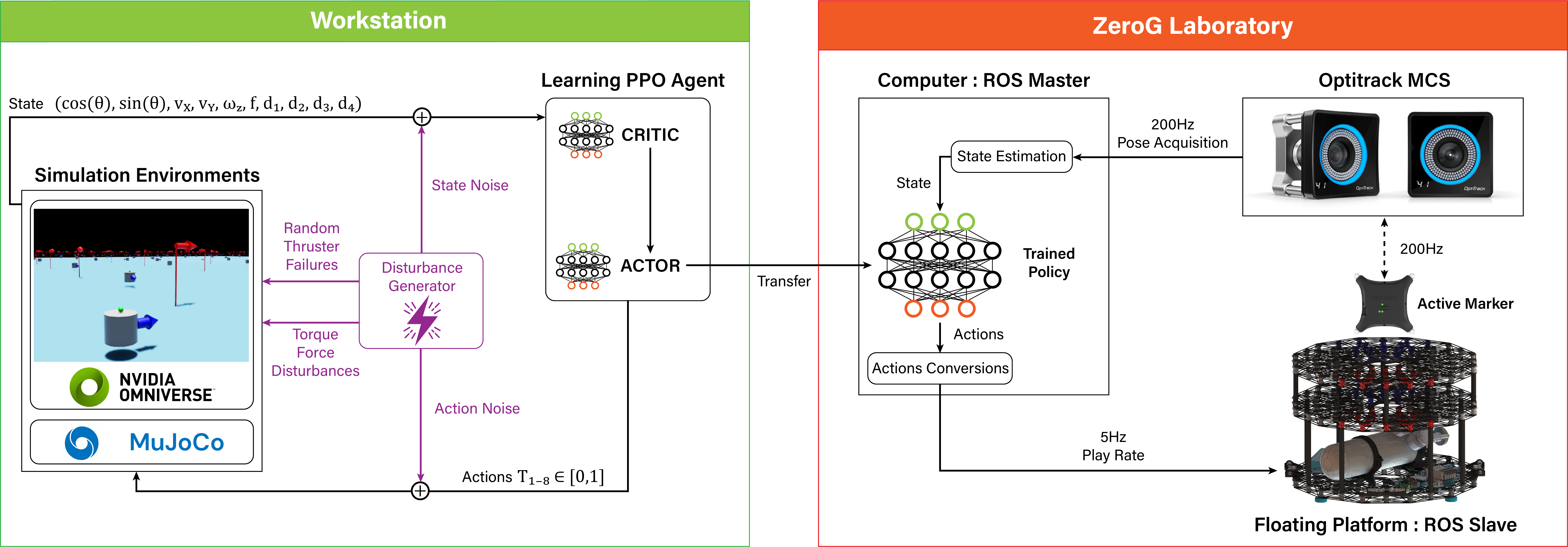}
    \caption{Framework Employed for Training and Evaluation: On the left, we depict the agent's interaction during both training and evaluation phases with the simulation environments, highlighting the incorporation of disturbances in the loop. On the right, we illustrate the deployment of the trained policy, while performing open-loop control on the real FP system.}
    \label{fig:framework}
      \vspace{-15pt}
\end{figure*}

\subsection{Training Procedure}
We reworked the PPO implementation from the RL Games library~\cite{rl-games2021} as the foundation of our training procedure.
This implementation utilizes GPU acceleration to vectorize observations and actions, enabling parallelization within the simulator by having both the simulation and the policy training residing on GPU.
Our agents are designed as actor-critic networks with two hidden layers, each consisting of 128 units. This makes them light and fast enough to be ran at high frequency on embedded devices. 
The hyper-parameters are listed in Table~\ref{Appendix:PPO} in the appendix. 
The agents train in their respective environments for 2000 epochs (approximately 130M steps). For more details about the network or the PPO configuration, we invite the reader to refer to the training configuration files available along with the code release at  \url{https://github.com/elharirymatteo/RANS}.

\subsection{Benchmark comparison with an Optimal Controller}
In this paper, we aim to provide a benchmark comparison between deep reinforcement learning and optimal control approaches, LQR in particular, for addressing the control problem of the floating platform in various scenarios.
Our objective is not to establish the superiority of one method over the other, but rather to gain insights into the strengths and weaknesses of each approach under different environmental conditions and task requirements.

An infinite horizon discrete-time LQR controller~\cite{stengel1994optimal} is used as a preliminary comparison with the DRL algorithm to control the FP. The LQR technique utilizes linearized dynamics to comprehensively model system behavior, providing optimal solutions with long-term stability while handling minor disturbances \cite{varghese2017optimal}. Their adaptability and relatively straightforward implementation have resulted in their adoption for numerous space applications \cite{khosravi2024design, muralidharan2023rendezvous, muralidharan2023ground}. In the case of a FP, the position, linear velocities, orientation quaternions, and angular velocities in the 2D plane are considered state variables of the system, $\mathbf{X}_i$. Since a FP operates at a relatively high frequency, a linearized system dynamics, defined as \eqref{eq:lqr_dynamics}
\looseness=-1
%
\begin{equation}\label{eq:lqr_dynamics}
    \mathbf{X}_{k+1} = \mathbf{A} \mathbf{X}_k + \mathbf{B} \mathbf{U}_k 
\end{equation}
is sufficient to predict the control output for incremental steps. The linearized system matrices, represented by $\mathbf{A}$ and $\mathbf{B}$, are the partial derivatives of the state vector at the final time step, denoted as $\mathbf{X}_{k+1}$, with respect to the current time step, $\mathbf{X}_{k}$, and the control input $\delta \mathbf{U}_k$, respectively. This computation leverages the central differencing technique, where the effects on the final states are evaluated in response to deliberate and minor perturbations applied to both the states and control inputs within the kinematic model simulated in Mujoco. To better account for the disturbances endured by the FP, the system matrices are updated at regular intervals. The LQR controller minimizes the cost function: 
\begin{equation*}\label{eq:cost_fn}
    J =  \sum_{k=0}^{\infty} \mathbf{X}_k^\text{T} \mathbf{Q} \mathbf{X}_k + \mathbf{U}_k^\text{T} \mathbf{R} \mathbf{U}_k
\end{equation*}
where $\mathbf{Q}$ and $\mathbf{R}$ are weighting matrices that penalize state errors and control outputs.
Minimizing the aforementioned cost function delivers an optimal control sequence given by:  
\begin{equation*}\label{eq:control_seq}
    \mathbf{U}_k = -\mathbf{K}\mathbf{X}_k
\end{equation*}
where $\mathbf{K}$ is the control feedback gain matrix defined by: 
\begin{equation*} \label{eq:feedback_gain}
    \mathbf{K}  = (\mathbf{R} + \mathbf{B}^\text{T} \mathbf{P} \mathbf{B})^{-1}\mathbf{B}^\text{T} \mathbf{P} \mathbf{A}
\end{equation*}
such that $\mathbf{P}$ is a positive definite matrix that is a solution for the Algebraic Riccati equation, as in: 
\begin{equation*} \label{eq:riccati}
    \mathbf{P}  =  \mathbf{Q} + \mathbf{A}^\text{T} \mathbf{P} \mathbf{A} - \mathbf{A}^\text{T} \mathbf{P} \mathbf{B}\mathbf{K}.
\end{equation*}
The optimal control output, $\mathbf{U}_k$, is an eight-dimensional array with real numbers. Note that the control outputs correspond to the actuation of the eight thrusters on the FP, hence an alternate vector $\mathbf{\bar{U}}_k$ is implemented that is a least squares solution to: 
\begin{equation*}\label{eq:least}
    \min \;\; || \; \mathbf{\bar{U}}_k - \mathbf{U}'_k \; ||^2 
\end{equation*}
where $\mathbf{U}'_k$ is the normalized vector of  $\mathbf{U}_k$ with values between 0 and 1. Moreover, for $\mathbf{\bar{U}}_k = [u_1, u_2, ..., u_8]$, each $u_i$ for $i \in \{1,2,...,8\}$ represents a binary variable, i.e., 
       $u_i \in \{0,1\}$  
signifying the actuation state of each thruster as either ``on'' or ``off''.
\looseness=-1


\subsection{Laboratory Experiment Setup} \label{subsec:lab-setup}
To validate our approach in a real-world scenario, we conducted experiments using the physical air bearings platform~\cite{fp_spacer} located within the ZeroG Laboratory at the University of Luxembourg. This specific platform floats on an epoxy floor, weighs 5.32 kg and measures 31~cm in radius and 45~cm in height.
It is equipped with a Raspberry Pi 4 for onboard control and communication.
The ZeroG Lab contains an Optitrack Motion Capture System (MCS) that precisely tracks the platform's pose at a frequency of 200~Hz.
We derive linear and angular velocities through simple forward differencing, that estimate the rate of change of positions and orientations over consecutive time-steps. Thanks to the relatively high accuracy of the MCS, and a reasonable averaging window, concerns about noise sensitivity are negligible. Our experimental setup maintains a connection between a laptop, the MCS, and the FP through a local network.
The laptop serves as the ROS (Robot Operating System) master node on the network, subscribing to the Optitrack node to acquire pose data and publishing the actions of the trained agents at a rate of 5~Hz.
This action frequency is deliberately constrained to prevent damage to the solenoid valves controlling the thrusters on the floating platform.
Figure \ref{fig:framework} illustrates the key components interacting during the simulated training and validation phase (on the left), and those interacting during the closed-loop control tests of the real FP system in the Lab (on the right).


\section{Experimental Setup}

Our experiments encompass both numerical simulation-based evaluations and real-world 
validations.
For the evaluation, each trained policy was tested across a diverse set of scenarios defined by various environmental conditions. 
\looseness=-1

\subsection{Performance Metrics}\label{subsec:metrics}
To evaluate the performance of the pose task in numerical simulations, we record 9 metrics: The percentage of time the agent spends under a given distance threshold during a single trajectory. This measure is then averaged across all experiments. For instance, $\text{PT}_5$ denotes the percentage of time spent under $5~\text{cm}$, we also record this for $2~\text{cm}$ ($\text{PT}_2$) and $1~\text{cm}$ ($\text{PT}_1$).
This measure is also applied to the heading of the agents when performing the pose task. In this case, $\text{OT}_5$ is the percentage of time spent under $5~\text{degrees}$, this measure is also done for $2~\text{degrees}$ ($\text{OT}_2$), and $1~\text{degrees}$ ($\text{OT}_1$).
Finally, we also record the absolute average linear velocity (ALV) and absolute average angular velocities (AAV). These metrics are compiled per trajectory, and averaged on the whole of them. This enables us to estimate how dynamic the agent's movements are. Furthermore, we monitor the average number of actions used per step (AAS), to evaluate the efficiency of the policy.

To evaluate the pose task in the lab, we only use the position and orientation error,
since we do not have enough experiments to compile more complete statistics. However, we do 
provide complete trajectories to better understand the behavior of the RL agent and LQR controller.

Finally, for the velocity tracking, we chose to apply the controllers on a trajectory tracking task.
For that, we wrote a simple trajectory tracker, that generates a velocity vector to track, based on a sequence of points to follow.
This vector is computed by taking the closest point that intersect with a circle of radius $r$ centered around the system. This radius, is a look-ahead-distance which can be tuned to adjust the speed of the tracker.
The velocity is considered fixed for the whole of the trajectory, meaning that the instructed velocity is not reduced even if there are sharp corners.
This controller is then applied on 3 shapes, a circle, a square and a infinite.
For these trajectories, we measure the error in velocity, and the averaged trajectory tracking error.

\subsection{Real-World Experimental Validations}
To validate the real-world applicability of our simulation-trained control policies, we used the physical floating platform with the laboratory setup described in section~\ref{subsec:lab-setup} to perform a series of experiments. Each test run, for the same policy, initiated the FP from different initial conditions, namely position and orientation within the lab.

\section{Results}

Simulation-based experiments demonstrate the efficacy of the PPO-based approach in achieving the defined tasks. The agent exhibits rapid task completion, stability in control, and adaptation to various scenarios. Quantitative metrics and qualitative visualizations substantiate the agent's high-performance capabilities.

\begin{table*}[t!]
    \centering
    \caption{
    Benchmark of the RL model and LQR controller under disturbances. For PT and OT, higher is better.
    For ALV, AAV, and AAS lower is better.
    Colors in the table indicate the drop in performance relative to their own ideal conditions:
    \textcolor{NavyBlue}{blue(0-20\%)}, \textcolor{Green}{green(20-40\%)}, \textcolor{YellowOrange}{yellow(40-60\%)}, \textcolor{OrangeRed}{red(60-80\%)}, \textcolor{RedViolet}{purple(80-100\%)}.
    The parameters of the dynamics of the LQR are tuned without noise or disturbances enabled.
    }
\resizebox{\textwidth}{!}{
\begin{tabular}{|l|l|cccc|ccccccccc|}
\toprule
\multirow{1}{*}{\textbf{Conditions}}& \multirow{1}{*}{\textbf{Controllers}} & \multicolumn{4}{c|}{\textbf{Disturbances}} & \multicolumn{9}{c|}{\textbf{Metrics}} \\
&  & VN & UF & TD & RTF & PT5 & PT2 & PT1 & OT5 & OT2 & OT1 & ALV & AAV & AAS \\
&  & (m/s) & (N) & (N$\cdot$m) & (-) & (\%) & (\%) & (\%) & (\%) & (\%) & (\%) & (m/s) & (rad/s) & (-) \\
\midrule\hline
\multirow{2}{*}{Ideal} &RL&-&-&-&-&\textcolor{black}{64}&\textcolor{black}{34}&\textcolor{black}{6}&\textcolor{black}{94}&\textcolor{black}{89}&\textcolor{black}{73}&0.08&0.12&0.29 \\
&LQR&-&-&-&-&\textcolor{black}{73}&\textcolor{black}{41}&\textcolor{black}{17}&\textcolor{black}{27}&\textcolor{black}{11}&\textcolor{black}{5}&0.07&0.16&0.10 \\\hline\hline
\multirow{4}{*}{Velocity Noise} &RL&0.02&-&-&-&\textcolor{NavyBlue}{64}&\textcolor{NavyBlue}{30}&\textcolor{NavyBlue}{7}&\textcolor{NavyBlue}{94}&\textcolor{NavyBlue}{90}&\textcolor{NavyBlue}{72}&0.08&0.12&0.31 \\
&RL&0.04&-&-&-&\textcolor{NavyBlue}{61}&\textcolor{Green}{21}&\textcolor{NavyBlue}{6}&\textcolor{NavyBlue}{94}&\textcolor{NavyBlue}{89}&\textcolor{NavyBlue}{66}&0.09&0.13&0.31 \\\cline{2-15}
&LQR&0.02&-&-&-&\textcolor{Green}{53}&\textcolor{YellowOrange}{21}&\textcolor{OrangeRed}{6}&\textcolor{RedViolet}{4}&\textcolor{RedViolet}{1}&\textcolor{RedViolet}{0}&0.09&0.49&0.23 \\
&LQR&0.04&-&-&-&\textcolor{OrangeRed}{14}&\textcolor{RedViolet}{3}&\textcolor{RedViolet}{0}&\textcolor{RedViolet}{2}&\textcolor{RedViolet}{1}&\textcolor{RedViolet}{0}&0.15&0.56&0.29 \\\hline\hline
\multirow{2}{*}{Constant Torque}&RL&-&-&0.05&-&\textcolor{NavyBlue}{63}&\textcolor{Green}{24}&\textcolor{YellowOrange}{2}&\textcolor{NavyBlue}{94}&\textcolor{NavyBlue}{86}&\textcolor{NavyBlue}{61}&0.08&0.12&0.35 \\\cline{2-15}
&LQR&-&-&0.05&-&\textcolor{Green}{57}&\textcolor{YellowOrange}{20}&\textcolor{OrangeRed}{6}&\textcolor{RedViolet}{3}&\textcolor{RedViolet}{1}&\textcolor{RedViolet}{0}&0.07&0.43&0.35 \\\hline\hline
\multirow{4}{*}{Constant Force}&RL&-&0.20&-&-&\textcolor{NavyBlue}{63}&\textcolor{NavyBlue}{29}&\textcolor{NavyBlue}{7}&\textcolor{NavyBlue}{94}&\textcolor{NavyBlue}{90}&\textcolor{NavyBlue}{74}&0.09&0.12&0.30 \\
&RL&-&0.40&-&-&\textcolor{NavyBlue}{52}&\textcolor{YellowOrange}{19}&\textcolor{NavyBlue}{5}&\textcolor{NavyBlue}{94}&\textcolor{NavyBlue}{89}&\textcolor{NavyBlue}{72}&0.09&0.12&0.31 \\\cline{2-15}
&LQR&-&0.20&-&-&\textcolor{NavyBlue}{66}&\textcolor{YellowOrange}{17}&\textcolor{OrangeRed}{4}&\textcolor{NavyBlue}{28}&\textcolor{NavyBlue}{12}&\textcolor{NavyBlue}{6}&0.07&0.15&0.12 \\
&LQR&-&0.40&-&-&\textcolor{OrangeRed}{23}&\textcolor{RedViolet}{0}&\textcolor{RedViolet}{0}&\textcolor{NavyBlue}{30}&\textcolor{NavyBlue}{13}&\textcolor{NavyBlue}{6}&0.08&0.16&0.15 \\\hline\hline
\multirow{2}{*}{Constant Force \& Torque}&RL&-&0.20&0.05&-&\textcolor{NavyBlue}{62}&\textcolor{Green}{24}&\textcolor{NavyBlue}{5}&\textcolor{NavyBlue}{94}&\textcolor{NavyBlue}{86}&\textcolor{NavyBlue}{61}&0.08&0.12&0.35 \\\cline{2-15}
&LQR&-&0.20&0.05&-&\textcolor{RedViolet}{13}&\textcolor{RedViolet}{2}&\textcolor{RedViolet}{0}&\textcolor{RedViolet}{3}&\textcolor{RedViolet}{1}&\textcolor{RedViolet}{0}&0.07&0.44&0.32 \\\hline\hline
\multirow{4}{*}{Thruster Failures}&RL&-&-&-&1&\textcolor{YellowOrange}{32}&\textcolor{YellowOrange}{15}&\textcolor{NavyBlue}{6}&\textcolor{Green}{70}&\textcolor{Green}{55}&\textcolor{YellowOrange}{36}&0.10&0.12&0.28 \\
&RL&-&-&-&2&\textcolor{OrangeRed}{15}&\textcolor{RedViolet}{6}&\textcolor{OrangeRed}{2}&\textcolor{YellowOrange}{45}&\textcolor{OrangeRed}{31}&\textcolor{OrangeRed}{20}&0.16&0.15&0.25 \\\cline{2-15}
&LQR&-&-&-&1&\textcolor{YellowOrange}{40}&\textcolor{YellowOrange}{17}&\textcolor{OrangeRed}{5}&\textcolor{Green}{20}&\textcolor{Green}{8}&\textcolor{Green}{4}&0.10&0.21&0.16 \\
&LQR&-&-&-&2&\textcolor{RedViolet}{12}&\textcolor{RedViolet}{4}&\textcolor{RedViolet}{1}&\textcolor{YellowOrange}{11}&\textcolor{OrangeRed}{4}&\textcolor{YellowOrange}{2}&0.14&0.28&0.22 \\\hline

\bottomrule
\end{tabular}
}
    \label{tab:my_label}
    \vspace{-12pt}
\end{table*}

\subsection{Numerical Simulation RL \& LQR}
In this section, we explore the behaviour of an RL agent trained to perform the ``go to pose'' task, and compare it to the LQR controller. We chose the ``go to pose'' task as it is a representative example, allowing us to assess the behaviour of different policies while controlling both the position and the orientation of the FP. 
To characterize the controllers' behaviors we expose them to a range of disturbances.
Neither the RL agents nor the LQR are specifically adapted to incorporate methods from robust RL or robust optimal control theory.
Yet, it is important to acknowledge that the RL agent was trained with some domain randomization to learn how to deal with force disturbances up to 0.25~N.
Both of them are evaluated in MuJoCo, with similarly randomized initial conditions.
In Table \ref{tab:my_label}, each line corresponds to an experiment, with various disturbances applied, and was compiled using 256 trajectories of 250 steps each.

First, the two test models are analyzed under ideal conditions with no disturbances.
From the PT metrics, it is evident that the LQR controller converges faster in position with better accuracy than the RL, owing to substantially longer durations where the LQR maintains a position error under 1~cm. 
We can also see that the RL controller first aligns its heading with the goal, as it spends almost all its time under the $5^{\circ}$ threshold.
This is a byproduct of its reward shaping, which incentivizes the convergence of the heading as much as the position. Hence, to score the maximum of points, aligning the heading first is a sound strategy as it is the easiest under ideal conditions.
Finally, AAS values show that the LQR is a lot more fuel efficient in these conditions, with 66\% less fuel used than the RL agent.

When considering the Velocity Noise (VN), it is observed that with the lowest noise level, the RL performances remain unchanged, while the LQR struggles, in particular with attitude control. With 0.04~m/s of noise, the performance of both controllers decreases. However, the RL controller is more resilient than the LQR controller to this kind of disturbance, even though it was not trained for it. In the interest of brevity, we do not report action noise value in the table, as we found their effect to be negligible on both controllers.
\looseness=-1

Furthermore, when examining the Torque Disturbance (TD) of 0.05 N$\cdot$m, equivalent to 1/6-th of the total torque capacity of the platform, the performance of both controllers experiences a noticeable reduction, particularly for the LQR controller.
A similar pattern is observed with the force disturbance (UF), which would be equivalent to an uneven floor in the lab. In this case, starting by applying 0.2N of force on the platform (equivalent to 1/5th of its maximum thrust), results in the performance of both controllers being close to the ideal conditions, with a small performance drop of the LQR in fine positioning.
When doubling it (0.4 N), the RL policy remains close to its baseline, but the LQR performance decreases, making it unable to maintain positions under the 2.5 cm threshold. Similar behaviours are observed upon the addition of both force and torque disturbances.

Finally, the thruster failures impact the performance of both controllers in the same manner. With a single failed thruster, both controllers perform relatively well, but the addition of a second thruster failure impedes the controller's ability to drive the FP to its defined goals.

Overall, while the LQR controller demonstrates greater efficiency and precision in position control with our current tuning, it encounters challenges when subjected to the selected range of disturbances. In contrast, RL exhibits a lower degree of energy conservation but offers stronger resilience when subject to a wide range of disturbances. It is possible that with a different cost function, better tuning of its weights, and a robust optimal control approach, the LQR becomes adept with these disturbances. Similarly, the RL agent could be induced to learn more conservative policy that uses less actions throughout the episodes, via adequate reward shaping.
However, the RL agent is not using a robust RL approach either, and domain randomization was only applied on force disturbances up to 0.25N, which is less than the disturbances it can overcome.

\subsection{ZeroG Laboratory}

For experiments with the real FP system, we report tests using both the RL and LQR methods for the ``go to pose" task, and tests using the RL agent only for the ``track velocity" task. 

\subsubsection{Go to pose}
The controllers are run on the FP, which is connected to a constant air supply through a tether.
This tether applies some light unknown disturbances such as a small torque and force to the platform.
Moreover, the system velocities are derived from the optitrack system. The observed velocities include minor noise and small delays due to network communication. 

\begin{figure}
    \centering{
    \begin{tabular}{c}
        \includegraphics[width=\linewidth]{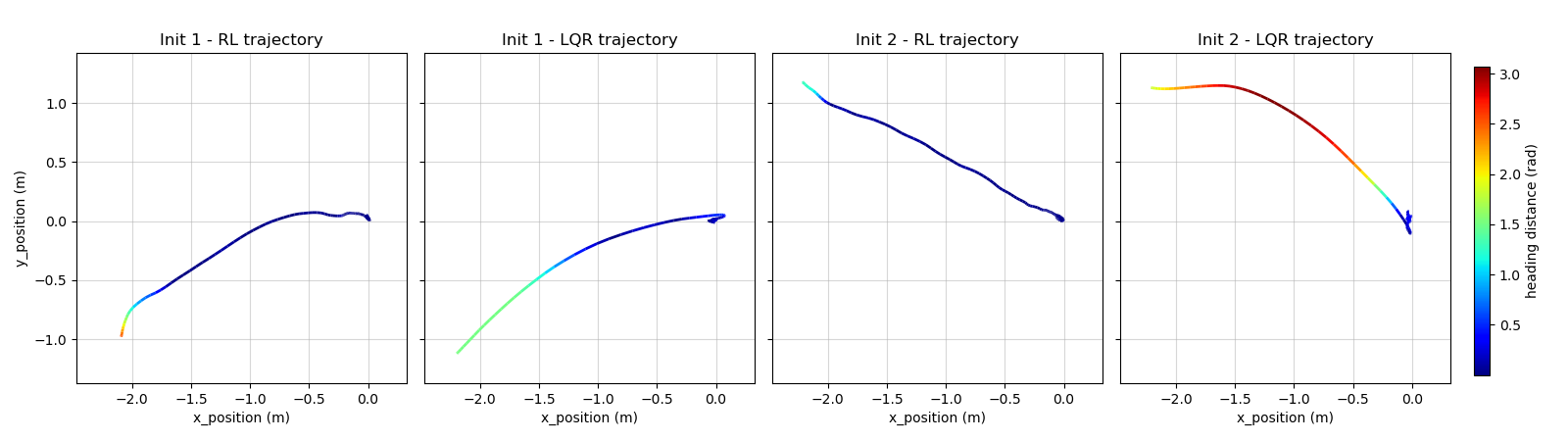}\\
        \includegraphics[width=\linewidth]{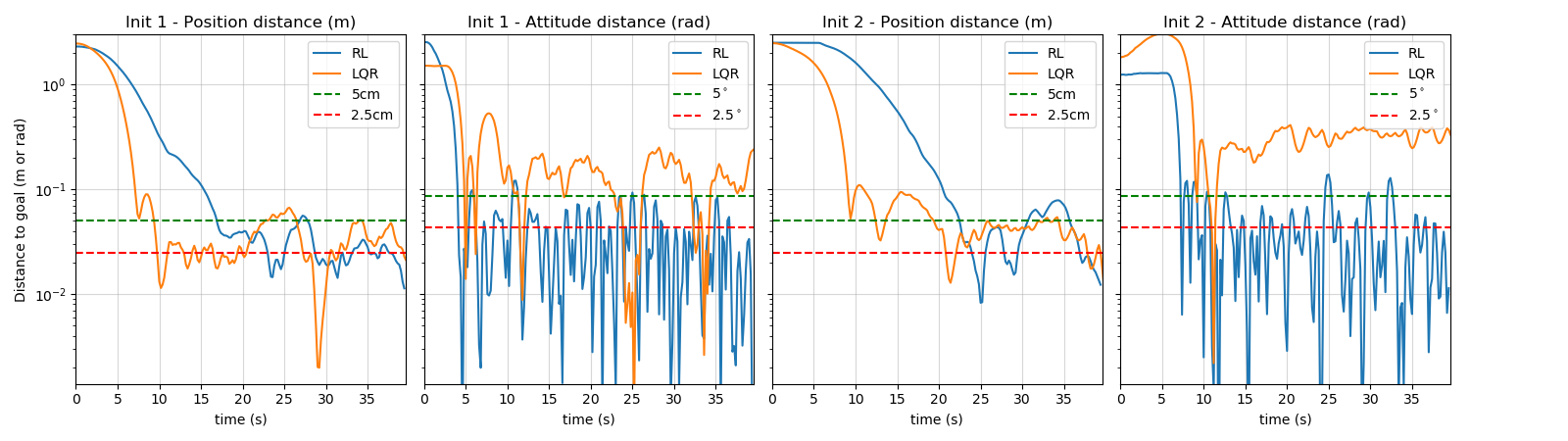}
    \end{tabular}}
    \caption{Comparison of the RL and LQR controller on two different initial poses in the ZeroG lab. Init 1 (resp. 2) denotes the first (resp. second) initial pose.
    For the trajectories, the scale of the y-axis is represented as a log value for better visualization. \looseness=-1}
    \label{fig:real_exp}
\end{figure}

Figure~\ref{fig:real_exp} illustrates the performance of each controller.
The first row shows the trajectories of the FP, and the second row shows the distance to the goal in position and orientation.
The first two columns have the rough same initial pose: Init1, while the two last share the same initial pose: Init2.

%

From the last row, it is evident that the LQR controller converges faster in position than the RL controller.
This aligns well with the behaviours observed in the simulation benchmark, with an LQR controller converging faster.
However, it is also apparent that the LQR solution exhibits a minor overshoot.
Such an observation is also in line with the simulation benchmark, as the uneven floor in the lab likely disrupts the LQR controller by applying a subtle constant force, preventing it from reaching its simulation baseline performance.
Looking at the top row, we can see that the LQR is also overshooting a bit.
Of course, the behaviour can be adjusted by modifying the weights associated with the importance of the error in position in the cost matrix.
It is also worth noting that the LQR controller is sensitive to the weights; smaller weights do not incentivize the FP motion toward the goal.
In comparison to the simulation, it was deemed necessary to alter the weights of the LQR controller to yield a more aggressive approach to achieve satisfying performances.
As for the RL agent, it is noticeable that the FP initially aligns its heading and then gradually converges toward the goal.
Consistently with the results from the simulation, the RL controller is significantly more accurate in terms of heading while achieving a position accuracy similar to that of the LQR controller.
Overall, both controllers performed well in the lab, reaching their expected performances.

\begin{figure}
    \centering
    \includegraphics[width=\linewidth]{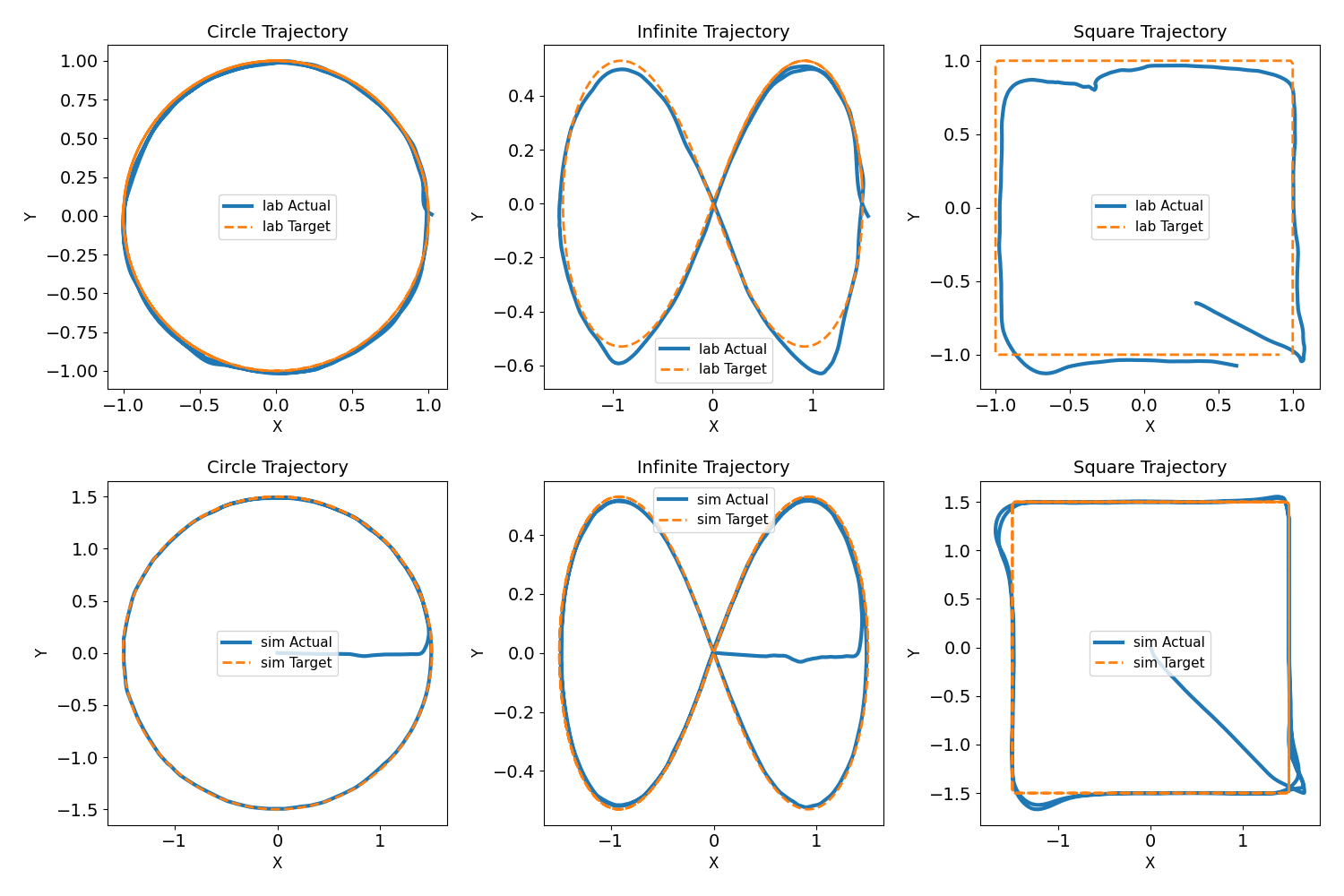}
    \caption{RL agent performing velocity tracking in simulated (bottom) and lab (top) environments.}
    \label{fig:track_vel_lab_vs_sim}
      \vspace{-10pt}
\end{figure}
\subsubsection{Track velocity} 

In the tests performed for this task in the lab, the objective is to assess the simulation-trained policy ability to adhere to a set of predetermined target velocities. Since the LQR model relies on both position and velocity states as input, while the RL agent only requires velocity, we opted to present the RL policy results for this specific task. Both numerical-simulation and lab tests are displayed to validate the sim-to-real transfer.

Similar to the ``go to pose'' experiments, the FP was subjected to un-modeled disturbances affecting both linear and angular motion. An additional challenge in these tests was the accurate estimation of velocities, affected by slight measurement noise and communication delays.
The pre-generated trajectories to be tracked by the policy were designed to test the FP's response accuracy and agility.

Figure~\ref{fig:track_vel_lab_vs_sim} illustrates the target trajectory and the FP's position for the circle, square and infinite shapes.
It is clearly visible that the hardest task was to follow a squared-shaped trajectory.
This is due to the sharp turns that require precise maneuvering and acceleration adjustments, which could be induced by reducing the look-ahead-distance and target velocity of the tracking when close to corners.
The performance metric used is the linear velocity error $e_v$ expressed as $\mu \pm \sigma$, where $\mu$ is the mean and $\sigma$ is the standard deviation during the test duration.
Table~\ref{tab:velocity_errors_env_comparison} reveals that the lab environment generally presents higher velocity errors compared to the simulation environment, particularly notable in the square shape with a lab error of 0.07 $\pm$ 0.05 m/s versus a sim error of 0.05 $\pm$ 0.08 m/s,  the difficulty of real-world transfer.
For the infinite trajectory, we observed a slight overshoot in the path's lower regions, caused by the irregularities in the epoxy floor, which are significant in that area of the laboratory, affecting the FP's motion.
This can also be seen on the square, and to less of a degree on the circle.
In our case, there is a slope pulling free-floating objects towards negative y.

\begin{table}[t!]
\centering
\begin{tabular}{lcc}
\hline
\textbf{Shape} & \textbf{Lab Error} ($\mu \pm \sigma$) [m/s] & \textbf{Sim Error} ($\mu \pm \sigma$) [m/s] \\ \hline
circle & 0.03 $\pm$ 0.02 & 0.01 $\pm$ 0.01 \\
infinite & 0.04 $\pm$ 0.03 & 0.01 $\pm$ 0.01 \\
square & 0.07 $\pm$ 0.05 & 0.05 $\pm$ 0.08 \\
\hline
\end{tabular}
\caption{Comparison of Velocity Errors Between Lab and Sim Environments for the track velocity task. All the trajectories are tracked at 0.2~m/s.} 
\label{tab:velocity_errors_env_comparison}
\vspace{-18pt}
\end{table}

\section{Conclusions}
This study presents a robust framework for the 2D control of a FP using deep reinforcement learning techniques. The enhancements to our simulator extend its applicability, enabling the training of agents for complex tasks. The demonstrated effectiveness of the PPO algorithm, with seamless transfer behavior from the simulation to the FP system, highlights its potential for autonomous navigation in space. Furthermore, our comparative analysis with traditional optimal control algorithms reveals the superior performance of our method in the presence of unpredictable stochastic disturbances. 
In the future, we aim to explore the efficacy of complex architectures such as LSTMs or Transformers to see if they can cope better with a greater number of disturbances and in particular actuators failure.
Additionally, we aim to broaden our framework to train agents capable of walking and jumping in microgravity using a combination of leg motion and thrusters, expanding the set of available sensors and actuators.





\section*{APPENDIX}

\subsection{PPO Hyperparameters}
Table~\ref{Appendix:PPO} outlines the key parameters used in the adapted version of the Proximal Policy Optimization (PPO) algorithm for training our models.
\begin{table}[h]
\centering
\begin{tabular}{cc}
\hline
\textbf{Parameter} & \textbf{Value} \\
\hline
Algorithm & PPO \\
\hline
Network Type & Actor-Critic MLP \\
\hline
Separate Networks & True \\
\hline
MLP Units & [128, 128] \\
\hline
Activation Function & tanh \\
\hline
Initializer & Identity \\
\hline
Regularizer & None \\
\hline
Learning Rate & $1e-4$ \\
\hline
Gamma ($\gamma$) & 0.99 \\
 \hline
Tau ($\tau$) & 0.95 \\
\hline
Entropy Coefficient & 0.0 \\
\hline
Horizon Length & 16 \\
\hline
Minibatch Size & 8192 \\
\hline
Mini Epochs & 8 \\
\hline
Critic Coefficient & 0.5 \\
\hline
Gradient Clipping Norm & 1.0 \\
\hline
KL Threshold & 0.016\\
\hline
Critic Coefficient & 0.5\\
\hline
\end{tabular}
\caption{PPO Training Parameters}
\label{Appendix:PPO}
\end{table}

\subsection{LQR parameters}

Table~\ref{tab:lqr_parameters} summarizes the parameters of the Discrete LQR Controller used. The controller is made planar compatible, indicating a restriction to the 2D plane. 

\begin{table}[h]
\centering
\resizebox{\linewidth}{!}{
\begin{tabular}{ll}
\hline
\textbf{Parameter} & \textbf{Value} \\
\hline
Name & $LQR$ \\
\hline
Q (State cost matrix) & [0.0001, 1e-05, 100, 100, 1e-06, 1e-06, 1] \\
\hline
R (Control cost matrix) & [0.01, 0.01, 0.01, 0.01, 0.01, 0.01, 0.01, 0.01] \\
\hline
W (Disturbance weight matrix) & [0.1, 0.1, 0.1, 0.1, 0.1, 0.1, 0.1] \\
\hline
Make planar compatible & Yes \\
\hline
Control type & LQR \\
\hline
\end{tabular}
}
\caption{Parameters for the Discrete LQR Controller}
\label{tab:lqr_parameters}
\end{table}
\vspace{-3pt}
\small

\section*{ACKNOWLEDGMENT}
  Work supported by the European Union's Horizon 2020 research and innovation program under grant No. 101096487.



\newpage




\end{document}